\documentclass[10pt,twocolumn,letterpaper]{article}

\usepackage{cvpr}
\usepackage{times}
\usepackage{epsfig}
\usepackage{graphicx}
\usepackage{amsmath}
\usepackage{amssymb}
\usepackage{booktabs}

\usepackage[breaklinks=true,bookmarks=false]{hyperref}

\cvprfinalcopy % *** Uncomment this line for the final submission

\def\cvprPaperID{****} % *** Enter the CVPR Paper ID here

\begin{document}

%%%%%%%%% TITLE
% \title{The AVA-Kinetics Action Localization Dataset}
\title{The AVA-Kinetics Localized Human Actions Video Dataset}

% \and

\author{Ang Li$^1$ \hspace{3em}
Meghana Thotakuri$^1$ \hspace{3em}
David A. Ross$^2$\\
Jo\~ao Carreira$^1$ \hspace{3em}
Alexander Vostrikov$^1$\footnotemark \hspace{3em}
Andrew Zisserman$^{1,4}$ \\
\small{$^1$DeepMind \hspace{3em}
 $^2$Google Research \hspace{3em}
 $^4$VGG, Oxford}\\
 \small\url{{anglili, sreemeghana, dross, joaoluis, zisserman}@google.com}\\
 \small\url{alexander.vostrikov@gmail.com}
}

\maketitle
\makeatletter{\renewcommand*{\@makefnmark}{}
\footnotetext{$^*$Work was done at DeepMind. Now with Citadel.}\makeatother}

%%%%%%%%% ABSTRACT
\begin{abstract}
This paper describes the AVA-Kinetics localized human actions video dataset. The dataset is collected by annotating videos from the Kinetics-700 dataset using the AVA annotation protocol, and extending the original AVA dataset with these new AVA annotated Kinetics clips. The dataset contains over 230k clips annotated with the 80 AVA action classes for each of the humans in key-frames.  We describe the annotation process and provide statistics about the new dataset. We also include a baseline evaluation using the Video Action Transformer Network on the AVA-Kinetics dataset, demonstrating improved performance for action classification on the AVA test set. The dataset can be downloaded from {\small\url{https://research.google.com/ava/}}.
\end{abstract}

%%%%%%%%% BODY TEXT
\section{The AVA-Kinetics Dataset} % Or perhaps The AVA-Kinetics Dataset Extension

The Kinetics~\cite{carreira2019short,kay2017kinetics} dataset was created to support representation learning in videos by providing a large classification task where researchers can explore architectures and pre-train models that can be finetuned on a variety of downstream tasks, similar to what happened with ImageNet \cite{imagenet_cvpr09} and image representations.

Downstream tasks tend to have more detailed annotations and are hence more expensive to annotate at scale. The AVA dataset~\cite{gu2018ava} presents one influential example of a video task that is expensive to annotate -- instead of a single label per clip as in Kinetics, every single person in a subset of frames gets a set of labels. This task is also interesting because pre-training on Kinetics leads to small improvements -- the state-of-the-art is still just at 34\% average precision~\cite{feichtenhofer2019slowfast}. 

This motivated us to create a crossover of the two datasets. The AVA-Kinetics dataset builds upon the AVA and Kinetics-700 datasets by providing AVA-style human action and localization annotations for many of the Kinetics videos. In this section, we introduce the AVA and Kinetics datasets briefly and describe the AVA annotation procedure applied to build the AVA-Kinetics dataset. The statistics of the dataset are also provided and analyzed in Section \ref{sec:stats}.

\begin{figure}
    \centering
    \includegraphics[width=\linewidth]{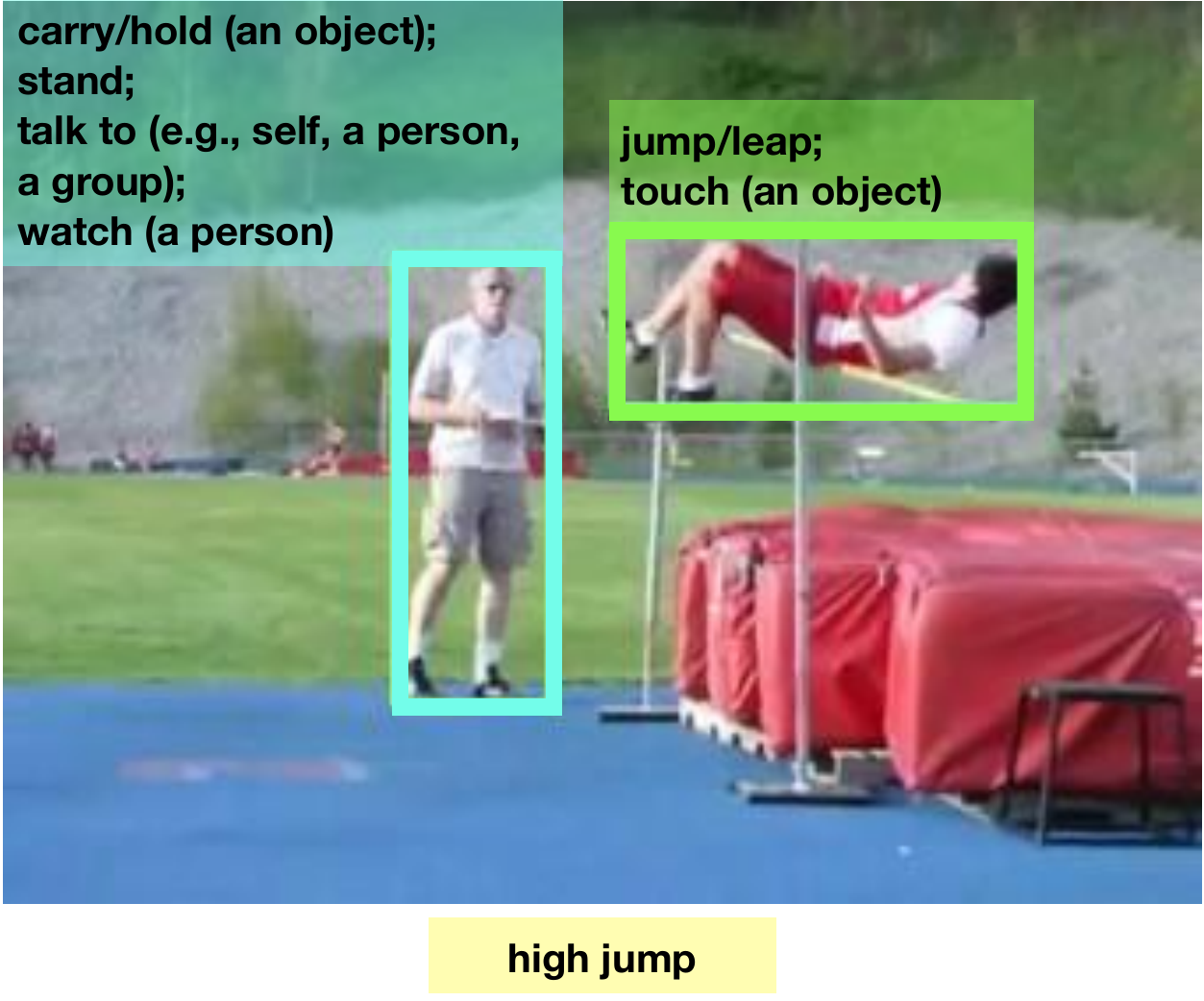}
    \caption{An example key-frame in the AVA-Kinetics dataset. The annotation contains AVA-style bounding boxes and their corresponding AVA labels. The key-frame is part of a Kinetics clip annotated with a clip-level label ``high jump''.}
    \label{fig:my_label}
\end{figure}
\subsection{Background}
The AVA dataset~\cite{gu2018ava} densely annotates 80 atomic visual actions in 430 15-minute movie clips. Human actions are annotated independently for each person in each video, on key-frames sampled once per second. The dataset also provides bounding boxes around each person and each person has all its co-occurring actions labeled (\eg, standing while talking) for a total of 1.6M labels. The Kinetics dataset is another large-scale curated video dataset for human action recognition covering a diverse range of human actions. Kinetics has progressed from Kinetics-400 to Kinetics-700 over the past few years. The Kinetics-700 dataset has 700 human action classes with at least 600 clips for each class, and a total of  around 650k video clips. For each class, each clip is from a different Internet video, lasts about 10s and has a single label describing the dominant action occurring in the video.

\subsection{Data Annotation Process}
The AVA-Kinetics dataset  extends the Kinetics dataset with AVA-style bounding boxes and atomic actions. A single frame is annotated for each Kinetics video, using a frame selection procedure described below. The  AVA annotation process is applied to a subset of the training data and to all video clips in the validation and testing sets from the Kinetics-700 dataset. The procedure to annotate bounding boxes for each Kinetics video clip was as follows:
\begin{enumerate}
    \item \textit{Person detection}: Apply a pre-trained Faster RCNN \cite{ren2015faster} person detector on each frame of the 10-second long video clips.
    \item \textit{Key-frame selection}: Choose the frame with the highest person detection confidence as the key-frame of each video clip, at least 1s away from the start/endpoint of the clip.
    \item \textit{Missing box annotation}: Human annotators  verify and annotate missing bounding boxes for the key-frame.
    \item \textit{Human action annotation}: Obtain a 2-second video clip centered on the key-frame. Multiple human raters (at least 3) then propose action labels corresponding to the persons in the key-frame bounding boxes.
    \item \textit{Human action verification}: Human raters assess all the proposed action labels for a final verification. Each label which is verified by a majority, of at least 2 of 3 raters, is retained.
\end{enumerate}

A difference from the original AVA dataset, is that only one key-frame is annotated for each Kinetics video clip.
The Kinetics training set is sampled for annotation as follows: first, we prioritize AVA classes with poor recognition performance by selecting a shortlist of 27 out of the original 80 AVA classes\footnote{The poorly performing AVA classes we prioritize are: swim, push (an object), give/serve (an object) to (a person), dress/put on clothing, watch (e.g., TV), throw, work on a computer, climb (e.g., a mountain), take (an object) from (a person), listen (e.g., to music), sing to (e.g., self, a person, a group), lift (a person), grab (a person), put down, cut, take a photo, pull (an object), enter, turn (e.g., a screwdriver), lift/pick up, point to (an object), push (another person), hit (an object), fall down, shoot, jump/leap, hand wave.} which have shown weak performance in the past literature. We hand-select 115 relevant action classes from the Kinetics dataset (listed in Appendix \ref{lbl:fully_annotated_classes}) by text matching to this shortlist, and the annotation pipeline is applied to all Kinetics videos containing those actions. Clips from the remaining Kinetics classes are sampled uniformly (we have not yet annotated all of them). Different from the training set, the Kinetics validation and test sets are both fully annotated.

\section{Data Statistics} \label{sec:stats}
We discuss in this section characteristics of the data distribution in AVA-Kinetics and compare it with the existing AVA dataset. The statistics of the dataset are given in Table~\ref{tab:stats} which shows the total number of unique frames and that of unique videos in these datasets. Kinetics dataset contains a large number of videos so it contributes a lot more unique videos to the AVA-Kinetics dataset. 

We show in this section the statistics of AVA v2.2 which contains 80 classes, however, in experiments, we follow the protocol of AVA challenge and only predict 60 classes.

\subsection{Sample distribution} 
\begin{figure*}[!t]
    \centering
    \includegraphics[width=\textwidth]{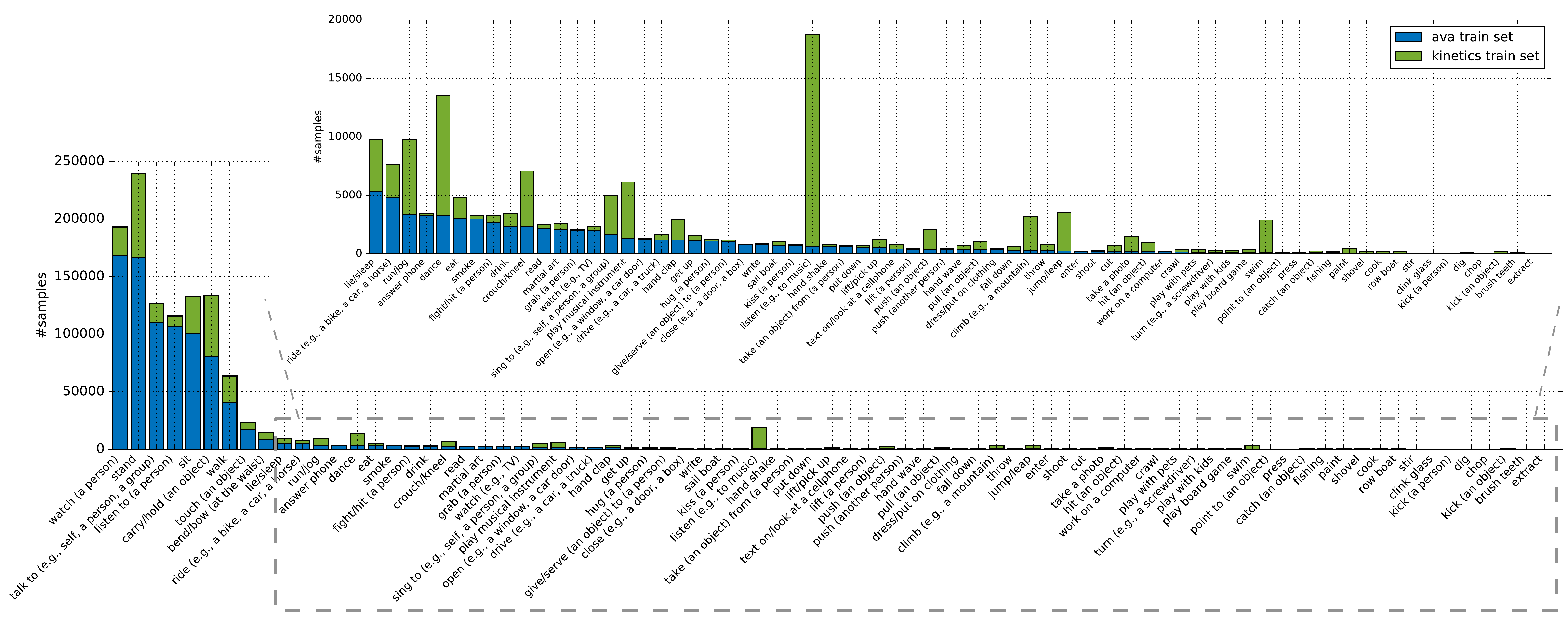}
    \caption{Number of samples per AVA class comparing the AVA train set (blue) and the Kinetics train set (green). The stacked bar shows the distribution of the AVA-Kinetics train set. The distribution is long-tailed. The tail part is zoomed in on the top right.}
    \label{fig:data_dist}
\end{figure*}

The number of samples for each class is given in Figure~\ref{fig:data_dist}. One sample corresponds to one bounding box with one action label. The class statistics in AVA and Kinetics  show similar trends, in particular both datasets exhibit a long-tailed sample distribution. However, we observe that Kinetics adds a substantial number of samples to most of the classes. Especially for the class ``listen to'', Kinetics introduces significantly more training samples.

\begin{table}[!t]
    \centering
        \setlength\tabcolsep{5pt}
    \caption{The number of annotated frames and the number of unique video clips in different splits for the AVA-Kinetics dataset. AVA-Kinetics data is a combination of AVA and Kinetics. While the numbers of annotated frames are roughly on par between AVA and Kinetics, Kinetics brings many more unique videos to the AVA-Kinetics dataset.}\vskip 0.5em
    \label{tab:stats}
    \resizebox{\linewidth}{!}{
    \begin{tabular}{crrr|rrr}
    \toprule
    & \multicolumn{3}{c|}{\bf \# unique frames} &\multicolumn{3}{c}{\bf \# unique videos}\\
         & \bf AVA & \bf Kinetics & \bf AVA-Kinetics& \bf AVA & \bf Kinetics & \bf AVA-Kinetics \\
    \midrule
    \bf Train     & 210,634  & 141,457 &352,091 & 235  & 141,240 & 141,475\\
    \bf Val & 57,371 &32,511 &89,882& 64 &32,465 & 32,529\\
    \bf Test &117,441 & 65,016& 182,457&131 & 64,771& 64,902\\
    \midrule
    \bf Total & 385,446 & 238,984 & 624,430 & 430 & 238,476 & 238,906 \\
    \bottomrule
    \end{tabular}}
\end{table}

\begin{figure*}[!t]
    \centering
    \includegraphics[width=\textwidth]{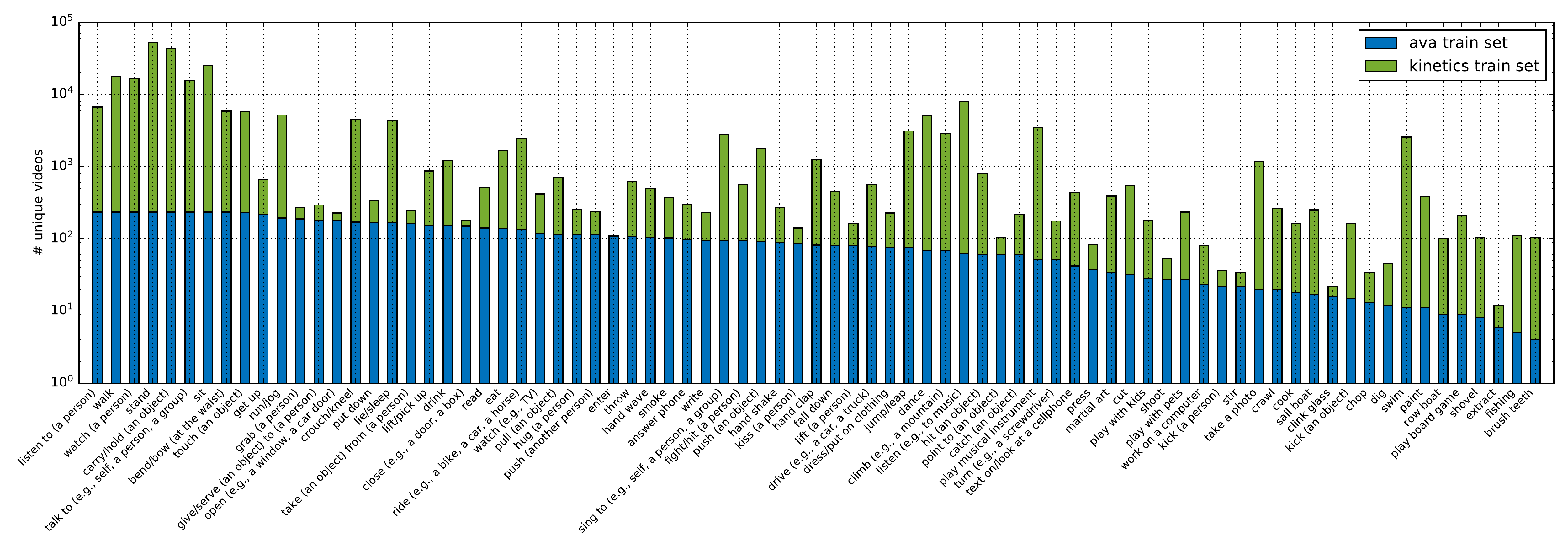}
    \caption{Number of unique videos per AVA class comparing the AVA train set (blue) and the Kinetics train set (green). The stacked bar shows the distribution of the AVA-Kinetics train set. The y-axis is in log-scale. The two datasets have different distribution in terms of videos. The AVA data has long videos while Kinetics has short video clips, hence the unique videos per class in AVA are much fewer.}
    \label{fig:data_dist_unique_video}
\end{figure*}

%a) Difference in number of examples per class between ava and ava-kinetics

\subsection{Video distribution}

Figure \ref{fig:data_dist_unique_video} shows the diversity of videos between the two datasets. The plot is in log scale and shows the property that Kinetics contains significantly more unique videos for each label. As stated before Kinetics is composed of many unique short videos, 10 second long, and a single key-frame is annotated with AVA labels. The AVA videos are much longer and are used to produce multiple clips for action localization. For AVA, the maximum number of unique videos per class is 235, while Kinetics has over 10,000 unique videos for the top classes. In addition, half of the classes in Kinetics have over 300 unique video clips.

\begin{table*}[!t]
    \centering
    \setlength\tabcolsep{10pt}
    \caption{A sampled set of Kinetics classes and the top related AVA classes ranked by Normalized pointwise mutual information (NPMI) on the AVA-Kinetics dataset. The NPMI score is shown next to each class label -- higher values indicate a stronger correlation.}
    \vskip 0.5em
    \label{tab:class-correlation}
    \resizebox{\textwidth}{!}{
    \begin{tabular}{ll}
    \toprule
    \bf Kinetics class&~\bf Top relevant AVA classes\\
    \midrule
putting on sari&~dress/put on clothing~{\footnotesize (0.73)};~stand~{\footnotesize (0.11)};~talk to (e.g., self, a person, a group)~{\footnotesize (0.04)};\\
talking on cell phone~~~~~~~~&~answer phone~{\footnotesize (0.74)};~text on/look at a cellphone~{\footnotesize (0.25)};~drive (e.g., a car, a truck)~{\footnotesize (0.16)};\\
slicing onion&~chop~{\footnotesize (0.70)};~cut~{\footnotesize (0.40)};~write~{\footnotesize (0.29)};\\
brushing teeth&~brush teeth~{\footnotesize (0.85)};~lift (a person)~{\footnotesize (0.19)};~text on/look at a cellphone~{\footnotesize (0.06)};\\
sailing&~sail boat~{\footnotesize (0.70)};~fishing~{\footnotesize (0.16)};~drive (e.g., a car, a truck)~{\footnotesize (0.10)};\\
canoeing or kayaking&~row boat~{\footnotesize (0.81)};~sail boat~{\footnotesize (0.43)};~fishing~{\footnotesize (0.32)};\\
casting fishing line&~fishing~{\footnotesize (0.73)};~turn (e.g., a screwdriver)~{\footnotesize (0.21)};~sail boat~{\footnotesize (0.20)};\\
shoveling snow&~shovel~{\footnotesize (0.71)};~drive (e.g., a car, a truck)~{\footnotesize (0.17)};~bend/bow (at the waist)~{\footnotesize (0.15)};\\
using a paint roller&~paint~{\footnotesize (0.76)};~give/serve (an object) to (a person)~{\footnotesize (0.09)};~talk to (e.g., self, a person, a group)~{\footnotesize (0.07)};\\
sipping cup&~drink~{\footnotesize (0.70)};~sit~{\footnotesize (0.22)};~lie/sleep~{\footnotesize (0.20)};\\
smoking hookah&~smoke~{\footnotesize (0.72)};~sit~{\footnotesize (0.14)};~answer phone~{\footnotesize (0.11)};\\
crawling baby&~crawl~{\footnotesize (0.80)};~lie/sleep~{\footnotesize (0.27)};~play with kids~{\footnotesize (0.19)};\\
dancing gangnam style&~dance~{\footnotesize (0.34)};~listen (e.g., to music)~{\footnotesize (0.28)};~watch (e.g., TV)~{\footnotesize (0.21)};\\
waiting in line&~text on/look at a cellphone~{\footnotesize (0.16)};~answer phone~{\footnotesize (0.15)};~stand~{\footnotesize (0.10)};\\
passing soccer ball&~kick (an object)~{\footnotesize (0.55)};~run/jog~{\footnotesize (0.34)};~give/serve (an object) to (a person)~{\footnotesize (0.14)};\\
\bottomrule%
    \end{tabular}}
\end{table*}

\begin{table*}[!t]
    \centering
    \setlength\tabcolsep{10pt}
    \caption{The least frequent AVA classes and their top relevant Kinetics classes ranked by Normalized pointwise mutual information (NPMI) on the AVA-Kinetics dataset. The NPMI score is shown next to each class label -- higher values indicate a stronger correlation.}\vskip 0.5em
    \label{tab:ava-kinetics-class-correlation}
    \resizebox{\textwidth}{!}{
    \begin{tabular}{ll}
    \toprule
    \bf AVA class&~\bf Top relevant Kinetics classes\\
    \midrule
close (e.g., a door, a box)&~closing door~{\footnotesize (0.69)};~opening refrigerator~{\footnotesize (0.54)};~shredding paper~{\footnotesize (0.42)};~~~~~~~~~~~~~~~~~~~~~~~~~~~~~~~~~~~~~~~~~~~~\\
chop&~slicing onion~{\footnotesize (0.70)};~chopping meat~{\footnotesize (0.57)};~making sushi~{\footnotesize (0.41)};\\
grab (a person)&~arresting~{\footnotesize (0.40)};~waxing armpits~{\footnotesize (0.31)};~shaving legs~{\footnotesize (0.28)};\\
enter&~cleaning pool~{\footnotesize (0.52)};~person collecting garbage~{\footnotesize (0.45)};~entering church~{\footnotesize (0.32)};\\
lift (a person)&~yoga~{\footnotesize (0.41)};~hugging baby~{\footnotesize (0.36)};~cracking back~{\footnotesize (0.33)};\\
shoot&~playing laser tag~{\footnotesize (0.43)};~playing paintball~{\footnotesize (0.42)};~lighting fire~{\footnotesize (0.38)};\\
kiss (a person)&~kissing~{\footnotesize (0.69)};~checking watch~{\footnotesize (0.49)};~marriage proposal~{\footnotesize (0.37)};\\
clink glass&~drinking shots~{\footnotesize (0.55)};~watching tv~{\footnotesize (0.30)};~falling off chair~{\footnotesize (0.28)};\\
point to (an object)&~presenting weather forecast~{\footnotesize (0.44)};~photocopying~{\footnotesize (0.40)};~playing slot machine~{\footnotesize (0.33)};\\
extract&~milking goat~{\footnotesize (0.56)};~shucking oysters~{\footnotesize (0.50)};~milking cow~{\footnotesize (0.49)};\\
kick (a person)&~drop kicking~{\footnotesize (0.52)};~high kick~{\footnotesize (0.40)};~cutting nails~{\footnotesize (0.38)};\\
dig&~clam digging~{\footnotesize (0.58)};~digging~{\footnotesize (0.45)};~planting trees~{\footnotesize (0.44)};\\
work on a computer&~leatherworking~{\footnotesize (0.43)};~coloring in~{\footnotesize (0.39)};~changing oil~{\footnotesize (0.39)};\\
press&~pulling espresso shot~{\footnotesize (0.51)};~milking cow~{\footnotesize (0.49)};~milking goat~{\footnotesize (0.48)};\\
stir&~scrambling eggs~{\footnotesize (0.57)};~making tea~{\footnotesize (0.54)};~making slime~{\footnotesize (0.49)};\\
\bottomrule
    \end{tabular}}
\end{table*}

\subsection{Correlation between Kinetics and AVA classes} Since we have annotations for both AVA classes and Kinetics classes for the Kinetics video, we compute the Normalized pointwise mutual information (NPMI) of the two types of classes. NPMI is defined as
\begin{align}
\textit{NPMI}(x, y) = \frac{\log p(x)+\log p(y)}{\log p(x,y)}-1
\end{align}
where $p(x)$ is the frequency of class $x$. $\textit{NPMI}(x,y)$ is a real value between -1 and 1. $\textit{NPMI}(\cdot,\cdot)=1$ means the two classes are highly correlated while $\textit{NPMI}(\cdot,\cdot)=-1$ means the two labels never co-occur.

We further rank the AVA classes for each Kinetics class according to their NPMI scores. We sample a few examples in Table \ref{tab:class-correlation} which also shows the corresponding NPMI scores.
% Kinetics classes usually describe certain types of activities while AVA labels describe more atomic human actions. For example, 
The Kinetics class ``dancing gangnam style'' is highly related to the AVA classes ``dance'', ``listen (e.g., to music)'' and ``watch (e.g., TV)''. While ``dance'' is directly related, ``listen'' and ``watch'' could also happen when someone is watching another person dancing gangnam style.

We show in Table \ref{tab:ava-kinetics-class-correlation} the reversed correlation, \ie, top Kinetics classes related to a certain AVA class. We pick a list of AVA classes with the least occurrence frequency and show their top relevant Kinetics class labels. The class ``stir'' is related to ``scrambling eggs'', ``making tea'' and ``'making slime''. The three cooking activities all requires a stirring action in practice.

\subsection{Person distribution} 
\begin{figure}[!t]
    \centering
    \includegraphics[width=\linewidth]{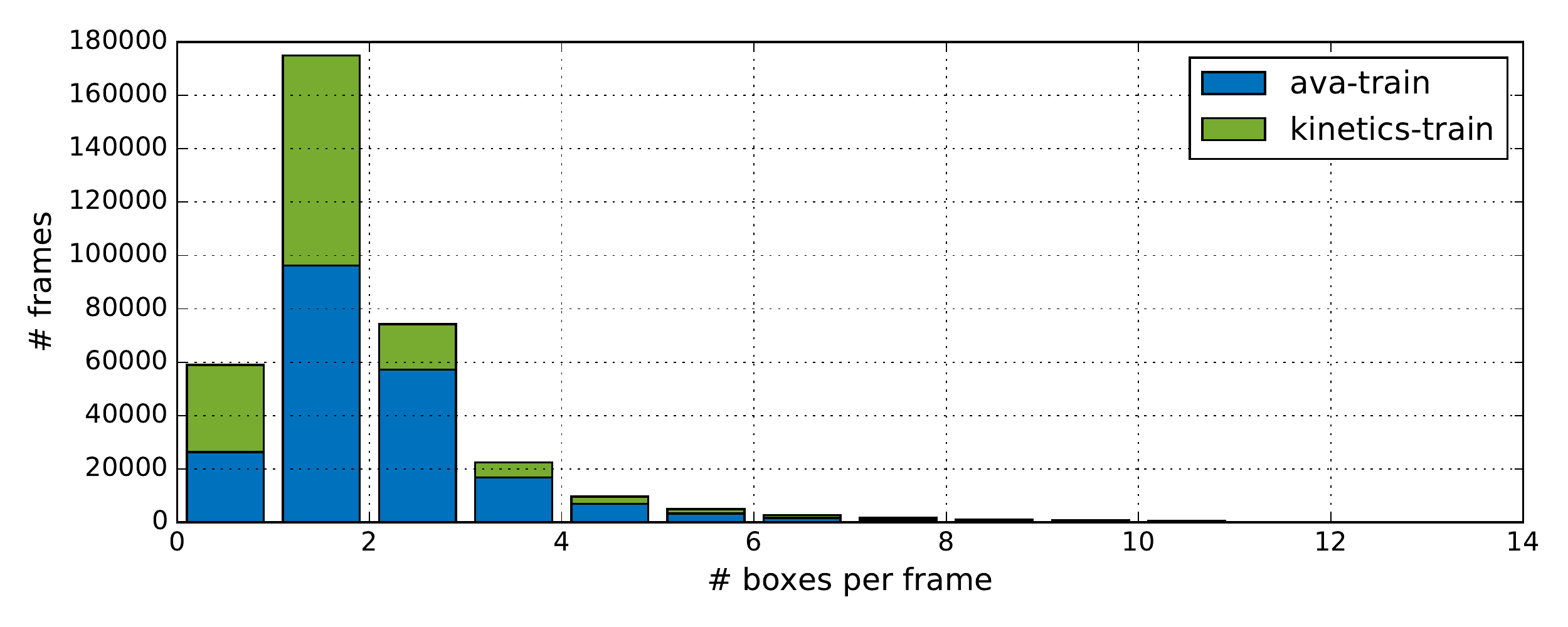}
    \caption{Number of bounding boxes per frame comparing the AVA train set (blue) and the Kinetics train set (green). The stacked bar shows the distribution of the AVA-Kinetics train set. There is a substantial number of frames with no person detected. The majority
of the key-frames contains only one bounding box.}
    \label{fig:nbox_per_frame}
\end{figure}

Figure \ref{fig:nbox_per_frame} shows the number of person bounding boxes per frame in both AVA and the Kinetics dataset. The distributions are roughly the same. We observe a substantial number of frames with no person detected in both datasets. And the majority of the key-frames have only 1 person detected. Frames that contain more than 5 persons are very rare. The average number of boxes per frame in AVA is 1.5 and that of Kinetics is around 1.2.

\subsection{Person box size distribution} 
\begin{figure}[!t]
    \centering
    \includegraphics[width=\linewidth]{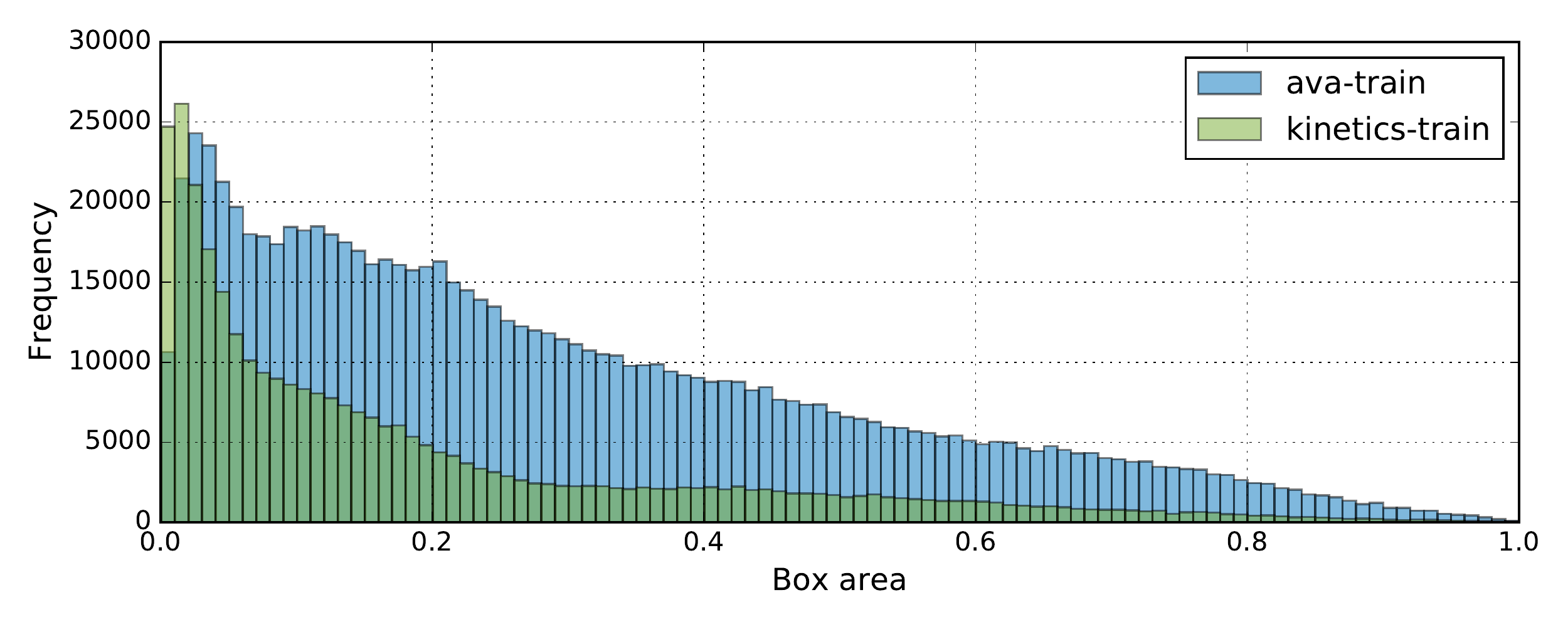}
    \caption{The distribution of the area of person bounding boxes in the AVA train set (blue) and the Kinetics train set (green). The area is normalized according to a 1x1 square. The peak area in AVA is around 0.02 while the peak in Kinetics is around 0.01. It is observed that Kinetics dataset produces significantly more smaller person bounding boxes.}
     \label{fig:box_area}
\end{figure}

We also study the size of the person bounding boxes in both datasets. Figure \ref{fig:box_area} shows the distribution of person bounding box areas. The area is normalized according to a 1x1 square image. We observe that the majority of the person boxes are small relative to the image. An interesting finding is that Kinetics videos tend to contain smaller person bounding boxes, compared to AVA videos.

%%%%%%

\section{Benchmarking Results}

We experiment with the new dataset using the Video Action Transformer Network~\cite{girdhar2019video} on top of ground truth person bounding boxes. Using ground truth boxes makes it possible to evaluate action classification accuracy separately from object detector failures. The original AVA paper~\cite{gu2018ava} reports 75\% mean AP person detection at 0.5 IoU using Faster RCNN with a ResNet-50 backbone; the Action Transformer~\cite{girdhar2019video} reports an  improvement on AVA score from 17.8\% to 29.1\% when using ground truth bounding boxes instead of automatically detected ones -- 11.2\% absolute. This is a significant improvement, but still small when considering the remaining 70\% left to get perfect performance. These results seem to suggest that plain action classification, of all things, is still very challenging in the multi-label case (Charades~\cite{sigurdsson2016hollywood} is another difficult multi-label action classification dataset, minus the spatial aspect). In addition to ground-truth based action classification, we also report results using a pre-trained person detector to propose boxes at the test time.

\subsection{Action Transformer: Short Recap}
The original Action Transformer~\cite{girdhar2019video} uses a 3D convolutional backbone to generate a spatiotemporal grid of features for each clip. In practice a video is divided into many clips, one around each key-frame. Using the spatiotemporal grid of features, a detector (e.g. the Region Proposal Network from Faster-RCNN) generates a number of object boxes for the middle frame. A positional embedding is added to the feature grid, which then gets ROI-pooled for each box and the resulting features are passed through a stack of transformers. The queries in the transformers are obtained by either averaging the spatio-temporal ROI-pooled features or using a more sophisticated merging operation that better preserves spatial configuration information -- for the purpose of ground truth box classification simpler averaging did better so we use that. The keys and values are directly derived from the original non-ROI-pooled grid of features. Overall the model tries to capture the relationship between each person and the whole scene.

Here we use a simplified model without any of the region proposal machinery. Instead we train on ground truth boxes then test on either ground truth boxes or on person detections provided by a more recent state-of-the-art detector~\cite{zhou2019objects}.

\subsection{Overall performance with groundtruth boxes} We show action classification performance with ground truth boxes in 9 different settings in Table \ref{tab:map_with_gtbox}, using three different training sets and three validation sets. AVA-Kinetics train/val is basically a combination of the corresponding AVA and Kinetics data. 

The table shows that the additional Kinetics training data leads to an improvement when evaluating on the original AVA validation set (+5.26 mAP). It also suggests that training on the Kinetics clips provides for more stable generalization: performance is close to an AVA-trained model on AVA validation, but the performance of the AVA-trained model is much lower on Kinetics validation clips. Finally training on the full AVA-Kinetics training set leads to the best performance on the full AVA-Kinetics validation set as would be expected.

\begin{table}[!t]
    \centering
    \caption{Performance of the Video Action Transformer on AVA action classification (mAP$\%$) using different train/val sets with given groundtruth bounding boxes.}\vskip 0.5em
    \label{tab:map_with_gtbox}
    \setlength\tabcolsep{5pt}
    \resizebox{\linewidth}{!}{
    \begin{tabular}{cccc}
    \toprule
         & \bf AVA & \bf Kinetics & \bf 
         AVA-Kinetics \\
         & val / test & val / test & val / test \\
         \midrule
    %\textbf{AVA} train     &  27.47 / 25.85 & 12.20 / 12.05& 18.44 / 17.61\\
    %\textbf{Kinetics} train & 22.09 / 20.91&  29.93 / 28.10 & 24.26 / 23.29\\
    %\textbf{AVA-Kinetics} train & \bf 32.73 / 31.30 &\bf  31.26 / 29.80 & \bf 31.42 / 30.42\\
    \textbf{AVA} train     &  27.47 / 25.85 & 16.08 / 16.02 & 24.26 / 23.47\\
    \textbf{Kinetics} train & 22.09 / 20.91 &  33.68 / 31.91 & 26.96 / 26.51 \\
    \textbf{AVA-Kinetics} train & \bf 32.74 / 31.30  &\bf  35.54 / 34.52 & \bf 35.98 / 35.56 \\
    \bottomrule
    \end{tabular}}
\end{table}

\begin{table}[!t]
    \centering
    \caption{Performance of the Video Action Transformer on AVA action classification (mAP$\%$) using different train/val sets with detected bounding boxes. The model is trained using just ground-truth bounding boxes.}\vskip 0.5em
    \label{tab:map_with_detbox}
    \setlength\tabcolsep{5pt}
    \resizebox{\linewidth}{!}{
    \begin{tabular}{cccc}
    \toprule
         & \bf AVA & \bf Kinetics & \bf AVA-Kinetics \\
         & val / test & val / test & val / test \\
         \midrule
    %\textbf{AVA} train     &  19.05 / 17.80 & 6.38 / 6.36 & 11.87 / 11.17\\
    %\textbf{Kinetics} train & 15.59 / 14.05 & 17.59 / 16.42& 15.59 / 14.73\\
    %\textbf{AVA-Kinetics} train & \bf 23.01 / 21.23 & \bf 18.38 / 17.61 & \bf 20.53 / 19.76\\
    \textbf{AVA} train    &  19.05 / 17.76  & 8.40 / 8.45  & 15.61 / 14.89\\
    \textbf{Kinetics} train & 15.59 /14.05 & 19.05 / 18.12 & 16.79 / 16.35 \\
    \textbf{AVA-Kinetics} train & \bf 23.01 / 21.23 & \bf 20.03 / 19.74  & \bf 22.99 / 22.70\\
    \bottomrule
    \end{tabular}}
\end{table}

\begin{figure*}[!t]
    \centering
    \includegraphics[width=\linewidth]{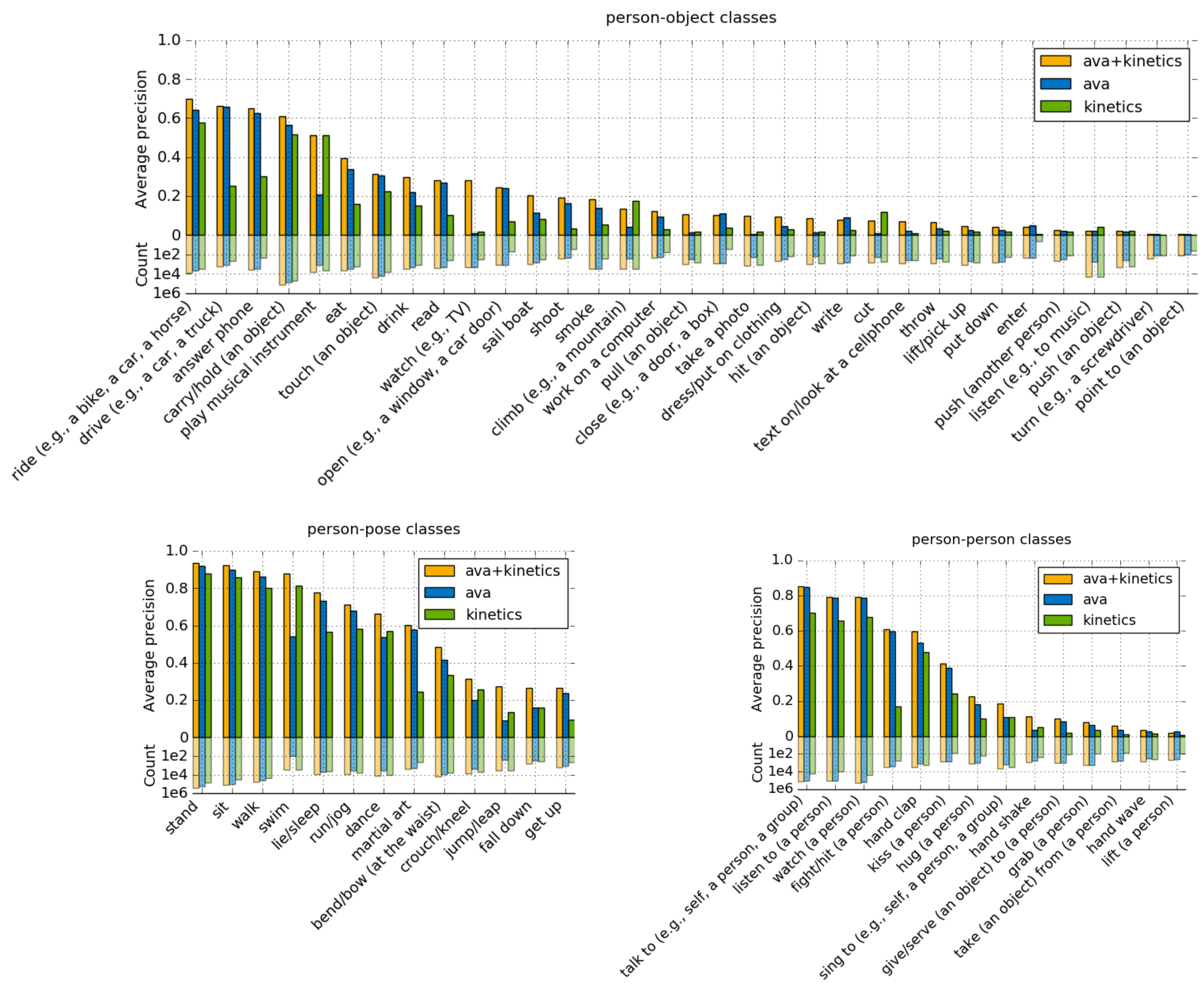}
    
    \caption{Per-class performance on AVA validation set for three action categories: person-person interaction, person-object interaction and person pose. The bars above zero-axis represent the average precision value while the transparent bars below the zero-axis represent the number of examples in the corresponding training datasets. The ground-truth boxes are used at both training and test time.}
    \label{fig:gt-per-class-map}
\end{figure*}

\begin{figure*}[!t]
    \centering
    \includegraphics[width=\linewidth]{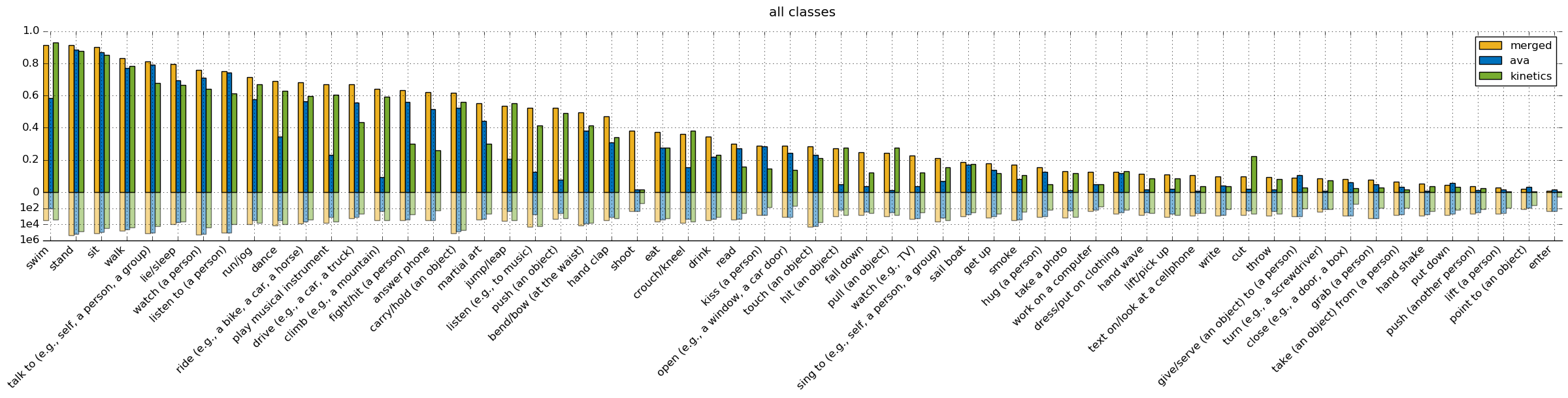}
    \caption{Per-Class Evaluation on AVA-Kinetics validation set. Compared models are trained on AVA, Kinetics and AVA-Kinetics (combined). The ground-truth boxes are used at both training and test time. The bars above zero-axis represent the average precision value while the transparent bars below the zero-axis represent the number of examples in the corresponding training datasets. }
    \label{fig:per-class-eval-on-avakinetics}
\end{figure*}

\subsection{Overall performance with detected boxes} Table~\ref{tab:map_with_detbox} shows the corresponding results when using automatically detected boxes at test time. We used the  CenterNet~\cite{zhou2019objects} person detector to generate a small set (about 2.5 boxes per key-frame) of confident detections that are fed to the video action transformer model. The object detector is trained on COCO \cite{coco} 2017 training set and achieves 43\% mAP on COCO 2017 validation set. Only those boxes detected with the ``person'' class are used as proposals for video action detection. For simplicity, the same model trained on ground truth boxes was employed -- the only change is at test time when detected boxes were used.

The results are mostly consistent with those evaluating on ground truth boxes, showing an improvement on AVA validation when training on the AVA-Kinetics training set (+3.96 mAP). Overall the values are slightly lower due to imperfect detection. The original action transformer model reports higher AVA-train to AVA-val mAP but it uses a more complex training procedure with a region proposal network and background negative example selection whereas  we use just ground truth boxes for training.

%\begin{table}[!t]
%    \centering
%    \caption{Performance of the Video Action Transformer on AVA action detection (mAP$\%$) %using different train/val sets with detected bounding boxes. The model is trained using %detected bounding boxes too.}\vskip 0.5em
%    \label{tab:map_with_gtbox}
%    \setlength\tabcolsep{5pt}
%    \resizebox{\linewidth}{!}{
%    \begin{tabular}{cccc}
%    \toprule
%         & AVA val & Kinetics val & AVA-Kinetics val \\
%         \midrule
%    AVA train     &  - & - & -\\
%    Kinetics train & - & - & -\\
%    AVA-Kinetics train & \bf - & - & -\\
%    \bottomrule
%    \end{tabular}}
%\end{table}

\subsection{Per-class performance} In addition to the overall comparison, we plot the per-class performance on same AVA validation set using the three  different training splits (AVA, Kinetics, AVA-Kinetics). The comparison is shown in Figure \ref{fig:gt-per-class-map} where all the AVA classes are divided into three categories: person-object interaction, person-pose and person-person interaction. The total number of training samples is also plotted in log-scale below the zero-axis in each sub-figure. It is observed that the overall performance on person-pose classes is relatively higher while person-object classes seem to be the most difficult cases to recognize. Notably, the performances on ``watch (e.g., TV)'', ``cut'', ``hand shake'', ``jump/leap'' and ``swim'' are significantly improved by using the new Kinetics data. A similar per-class evaluation on the full AVA-Kinetics validation set is shown in Figure \ref{fig:per-class-eval-on-avakinetics}.

\subsection{Performance improvement vs. data increase}
\begin{figure}[!t]
    \centering
    \includegraphics[width=\linewidth]{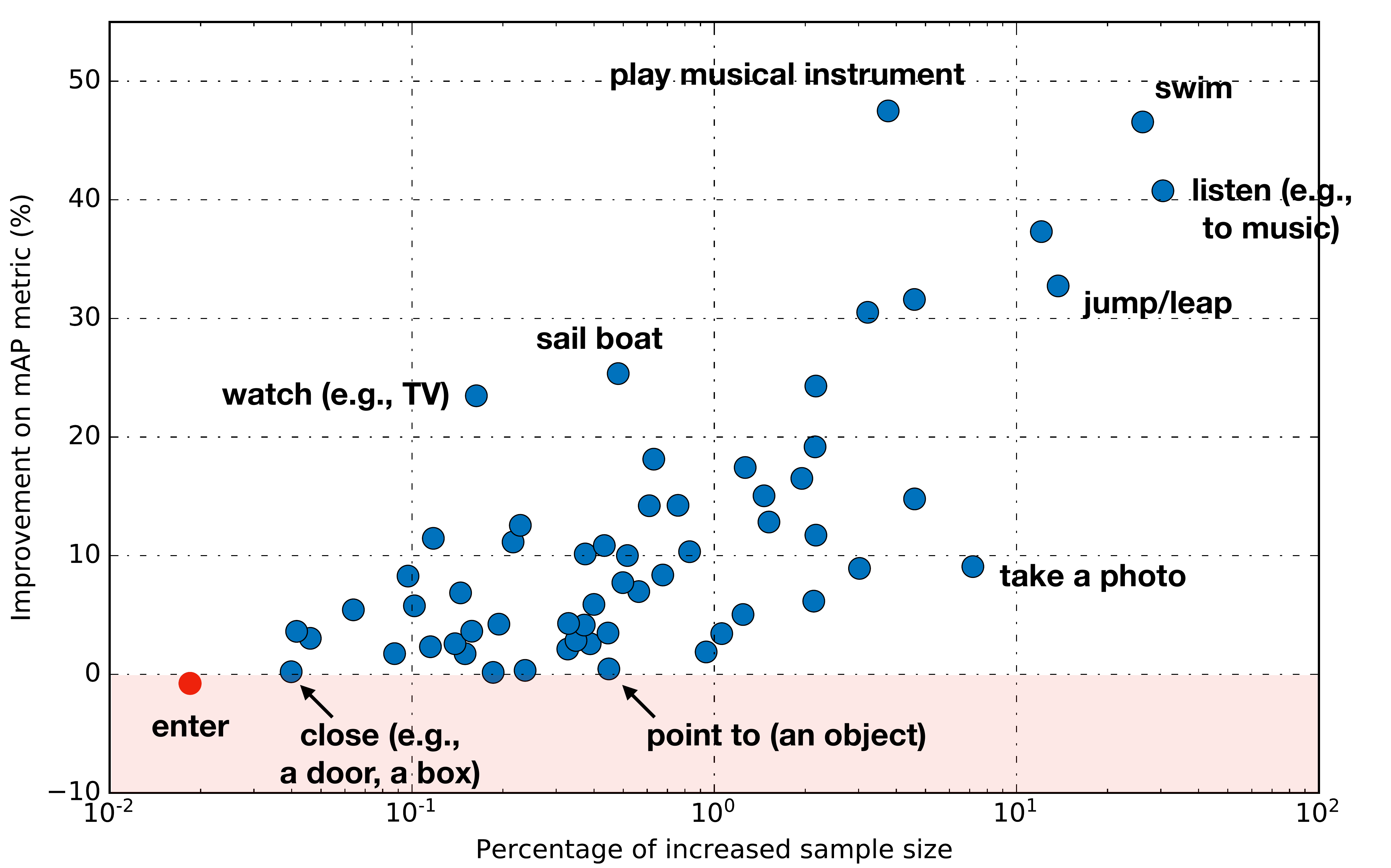}
    \caption{The improvement of mAP performance on AVA-Kinetics validation set. The x-axis represents the percentage of added Kinetics sample size w.r.t. the sample size in AVA training set. Each point corresponds to one class. Blue dots represent classes with improved mAP while the only red dot represent the class ``enter'' with decreased mAP. However, the mAP decrease ($-0.76\%$) is insignificant, just below zero. The results are obtained using the groundtruth boxes at test time.}
    \label{fig:per-class-improve}
\end{figure}

We study in Figure \ref{fig:per-class-improve} how the performance improvement is related to the increase of data. The $x$-axis is the percentage of increased sample size from AVA training set to AVA-Kinetics training set. This is basically the ratio of Kinetics sample size vs. AVA sample size on the same label class. The $y$-axis is the improvement of mAP, \ie, the mAP of a model trained on AVA-Kinetics set subtracts the mAP of a model trained on AVA set. Some of the class names are shown nearby the corresponding data points. The mAP metric is measured on the AVA-Kinetics validation set. We observe that only one class ``enter'' (red colored) has a performance drop ($0.76\%$ mAP lower). All other classes benefit from training on more Kinetics videos. Among the most improved classes are ``play musical instrument'', ``swim'' and ``'listen (e.g., to music)'' -- in these cases the number of training examples has more than doubled the original.

\section{Conclusions}

We have presented the AVA-Kinetics Localized Human Actions Video Dataset which is a crossover of the Kinetics and AVA datasets. AVA has detailed multi-label annotations for all people but restricted visual diversity at around 500 unique videos; Kinetics has a single label per video but a very wide visual diversity given its 600k unique videos. By annotating one frame in each Kinetics video with AVA boxes and labels, we obtain the AVA-Kinetics dataset with a richer training set and also a test set that better reflects true model generalization.

We believe this dataset can be a useful aid for research in a variety of settings in video. For example in multi-task learning (training on both AVA and Kinetics labels  vs individual ones) or in transfer learning (how does   pre-training on Kinetics then finetuning on AVA compare to directly training on AVA-Kinetics?).

\subsection*{Acknowledgements:} The collection of this dataset was funded by DeepMind and Google Research. The authors would like to thank Vighnesh Birodkar, Yu-hui Chen, Jonathan Huang, Vivek Rathod for their help on the object detection, Yeqing Li for his help on the action annotation pipeline, and Jean-Baptiste Alayrac for reviewing a paper draft.
We would also like to thank Rahul Sukthankar, Victor Gomes, Ellen Clancy and Chloe Rosenberg for their contributions to this work.

{\small
\bibliographystyle{ieee_fullname}
\bibliography{egbib}

\begin{thebibliography}{10}\itemsep=-1pt

\bibitem{carreira2019short}
Joao Carreira, Eric Noland, Chloe Hillier, and Andrew Zisserman.
\newblock A short note on the kinetics-700 human action dataset.
\newblock {\em arXiv preprint arXiv:1907.06987}, 2019.

\bibitem{imagenet_cvpr09}
J. Deng, W. Dong, R. Socher, L.-J. Li, K. Li, and L. Fei-Fei.
\newblock {ImageNet: A Large-Scale Hierarchical Image Database}.
\newblock In {\em CVPR09}, 2009.

\bibitem{feichtenhofer2019slowfast}
Christoph Feichtenhofer, Haoqi Fan, Jitendra Malik, and Kaiming He.
\newblock Slowfast networks for video recognition.
\newblock In {\em Proceedings of the IEEE International Conference on Computer
  Vision}, pages 6202--6211, 2019.

\bibitem{girdhar2019video}
Rohit Girdhar, Joao Carreira, Carl Doersch, and Andrew Zisserman.
\newblock Video action transformer network.
\newblock In {\em Proceedings of the IEEE Conference on Computer Vision and
  Pattern Recognition}, pages 244--253, 2019.

\bibitem{gu2018ava}
Chunhui Gu, Chen Sun, David~A Ross, Carl Vondrick, Caroline Pantofaru, Yeqing
  Li, Sudheendra Vijayanarasimhan, George Toderici, Susanna Ricco, Rahul
  Sukthankar, Cordelia Schmid, and Jitendra Malik.
\newblock Ava: A video dataset of spatio-temporally localized atomic visual
  actions.
\newblock In {\em Proceedings of the IEEE Conference on Computer Vision and
  Pattern Recognition}, pages 6047--6056, 2018.

\bibitem{kay2017kinetics}
Will Kay, Joao Carreira, Karen Simonyan, Brian Zhang, Chloe Hillier, Sudheendra
  Vijayanarasimhan, Fabio Viola, Tim Green, Trevor Back, Paul Natsev, Mustafa
  Suleyman, and Andrew Zisserman.
\newblock The kinetics human action video dataset.
\newblock {\em arXiv preprint arXiv:1705.06950}, 2017.

\bibitem{coco}
Tsung-Yi Lin, Michael Maire, Serge Belongie, James Hays, Pietro Perona, Deva
  Ramanan, Piotr Doll{\'a}r, and C.~Lawrence Zitnick.
\newblock Microsoft coco: Common objects in context.
\newblock In {\em European Conference on Computer Vision}, pages 740--755.
  Springer, 2014.

\bibitem{ren2015faster}
Shaoqing Ren, Kaiming He, Ross Girshick, and Jian Sun.
\newblock Faster r-cnn: Towards real-time object detection with region proposal
  networks, 2015.

\bibitem{sigurdsson2016hollywood}
Gunnar~A Sigurdsson, G{\"u}l Varol, Xiaolong Wang, Ali Farhadi, Ivan Laptev,
  and Abhinav Gupta.
\newblock Hollywood in homes: Crowdsourcing data collection for activity
  understanding.
\newblock In {\em European Conference on Computer Vision}, pages 510--526.
  Springer, 2016.

\bibitem{zhou2019objects}
Xingyi Zhou, Dequan Wang, and Philipp Kr{\"a}henb{\"u}hl.
\newblock Objects as points.
\newblock {\em arXiv preprint arXiv:1904.07850}, 2019.

\end{thebibliography}
}

\appendix
\section{Fully annotated Kinetics classes}
\label{lbl:fully_annotated_classes}
As mentioned above, we hand picked 115 Kinetics classes that are most related to the hardest AVA classes and annotated all of their training set examples. The full list is the following:

\begin{quotation}
swimming,
swimming backstroke,
swimming breast stroke,
swimming butterfly stroke,
swimming front crawl,
swimming with dolphins,
swimming with sharks,
scuba diving,
diving cliff,
helmet diving,
pushing car,
pushing cart,
pushing wheelbarrow,
pushing wheelchair,
giving or receiving award,
putting on sari,
putting on shoes,
watching tv,
card throwing,
catching or throwing baseball,
catching or throwing frisbee,
catching or throwing softball,
hammer throw,
javelin throw,
throwing axe,
throwing ball (not baseball or American football),
throwing discus,
throwing knife,
throwing snowballs,
throwing tantrum,
throwing water balloon,
assembling computer,
climbing a rope,
climbing ladder,
climbing tree,
ice climbing,
mountain climber (exercise),
rock climbing,
giving or receiving award,
marriage proposal,
headbanging,
listening with headphones,
silent disco,
beatboxing,
gospel signing in church,
singing,
karaoke,
carrying baby,
pull ups,
push up,
wrestling,
arresting,
arm wrestling,
wrestling,
shaking hands,
tango dancing,
clean and jerk,
dealing cards,
building lego,
card stacking,
sipping cup,
cutting watermelon,
cutting pineapple,
cutting orange,
cutting cake,
cutting apple,
taking photo,
pulling rope (game),
entering church,
using a wrench,
sharpening pencil,
using a paint roller,
deadlifting,
lifting hat,
snatch weight lifting,
stacking dice,
sign language interpreting,
pushing wheelchair,
tackling,
playing american football,
mosh pit dancing,
wrestling,
punching person (boxing),
hitting baseball,
punching bag,
golf chipping,
golf driving,
golf putting,
flint knapping,
falling off bike,
falling off chair,
faceplanting,
triple jump,
long jump,
high jump,
bungee jumping,
parkour,
gymnastics tumbling,
playing laser tag,
playing paintball,
triple jump,
high jump,
long jump,
jumping jacks,
bungee jumping,
springboard diving,
bouncing on trampoline,
gymnastics tumbling,
jumping sofa,
jumping into pool,
waving hand,
finger snapping,
drumming fingers,
playing hand clapping games,
pumping fist
\end{quotation}
\end{document}